\documentclass[10pt,twocolumn,letterpaper]{article}

\usepackage{cvpr}
\usepackage{times}
\usepackage{epsfig}
\usepackage{graphicx}
\usepackage{amsmath}
\usepackage{amssymb}

\usepackage{url}            % simple URL typesetting
\usepackage{booktabs}       % professional-quality tables
\usepackage{amsfonts}       % blackboard math symbols
\usepackage{nicefrac}       % compact symbols for 1/2, etc.
\usepackage{microtype}      % microtypography
\usepackage{multirow}
\usepackage{threeparttable}
\usepackage{caption}
\usepackage{subfigure}
\usepackage{wrapfig}
\usepackage{color}
\usepackage{diagbox}

\usepackage{appendix}
\cvprfinalcopy % *** Uncomment this line for the final submission

\def\cvprPaperID{****} % *** Enter the CVPR Paper ID here

\begin{document}

%%%%%%%%% TITLE
\title{Chinese Herbal Recognition based on Competitive Attentional Fusion of Multi-hierarchies Pyramid Features}

\author{Yingxue Xu,
        Guihua Wen \thanks{corresponding author},
        Yang Hu \thanks{equal contribution with Guihua Wen},
		Mingnan Luo,
        Dan Dai,
        Yishan Zhuang\\
School of Computer Science \& Engineering, South China University of Technology\\
Panyu, Guangzhou, Guangdong, China\\
\small
\{201530381885@mail., crghwen@, cssuperhy@mail., csluomingnan@mail., csdaidan@mail., cszhuangyishan@mail.\}scut.edu.cn
}

\maketitle
%\thispagestyle{empty}

%%%%%%%%% ABSTRACT
\begin{abstract}
Convolution neural netwotks (CNNs) are successfully applied in image recognition task. In this study, we explore the approach of automatic herbal recognition with CNNs and build the standard Chinese herbs datasets firstly. According to the characteristics of herbal images, we proposed the competitive attentional fusion pyramid networks to model the features of herbal image, which mdoels the relationship of feature maps from different levels, and re-weights multi-level channels with channel-wise attention mechanism. In this way, we can dynamically adjust the weight of feature maps from various layers, according to the visual characteristics of each herbal image. Moreover, we also introduce the spatial attention to recalibrate the misaligned features caused by sampling in features amalgamation. Extensive experiments are conducted on our proposed datasets and validate the superior performance of our proposed models. The Chinese herbs datasets will be released upon acceptance to facilitate the research of Chinese herbal recognition.
\end{abstract}

%%%%%%%%% BODY TEXT
\section{Introduction}
Deep convolution neural networks (CNNs) have achieved a grate success on in the field of image processing\cite{simonyan2014very,szegedy2015going,huang2017densely,Girshick2015Fast} and been applied on object detection and recognition\cite{he2016deep,liu2016ssd,redmon2016you} and get a better performance. As a kind of poor side effect, simple and noninvasive treatment, Chinese herbs are widely used in China and a number of Asian countries for healthcare\cite{yang2016impact,yang2015intelligent}. Therefore, there are wide application values and significance for recognizing Chinese herbs automatically. However, as far as we know, there is no research on this task and meanwhile it is difficult to train models for herbal recognition due to a lack of sufficient herbs data.\\
\indent
In this paper, we first propose a CNN model to deal with Chinese herbal recognition task, based on which we present a standard dataset for Chinese herbal recognition. Distinguishing from regular task of object recognition\cite{he2016deep,huang2017densely} and fine-grained image recognition\cite{fu2017look,Zheng2017Learning}, the former focus on distinguishing the outline and shape of object and the latter need more detailed features to identify so that they can classify with similar shape but different details. For Chinese Herbal recognition, we would be confronted with the above two cases: (a) some herbs are so distinguishing that they are easy to be classified with the shape features instead of detailed features. (b) some herbs with similar shapes usually need to be classified by more fine-grained features. The features extracted from convolution layers of different depth are rich in diversity that the features from earlier layers are more representational and from deeper layers are more abstract and contain more semantics in contrast\cite{liu2016ssd,Lin2016Feature}. According to the aforementioned challenges of herbal recognition, we choose Feature Pyramid Networks\cite{Lin2016Feature} (FPN) to merge features from different levels so that we can diversify image features overall to improve the performance of herbal recognition with CNNs.\\
\indent
Compared with the traditional FPN\cite{Lin2016Feature}, in this study, we first introduce channel-wise attention\cite{Hu2017Squeeze} in the process of fusing features from different levels. In this way, our models can dynamically adjust the weights of features from different levels, which makes it possible to adjust the extent of selecting features encoding from various levels adaptively. Furthermore, we also combine spatial attention\cite{Li2018Harmonious} to spatially recalibrate the misaligned features caused by serveral upsampling or downsampling operators during feed-forward propagation.\\
\indent
More importantly, the channel-wise and spatial attention are both improved in this paper as follows: (a) The original SE mechanism is limited on re-scaling the weights of features from the same layers, while the competitive attention proposed in this paper extends the modeling ranges of channel-wise attention, as same as spatial attention, and explicitly model the competitive channel dependencies between spatial and semantic informations in the process of fusion on various levels. (b) The feature maps from bottom-up pathway, which are abundant in more spatial informations to provide references for the misaligned and spatially coarser features from top-down pathway, are introduced into the process of spatial attentional modeling to recalibrate the misaligned features. Based on the above improvement of attention according to our specific structures and tasks, we can jointly model channel relationship of various levels and channel dependencies between spatial and semantic information flows, as well as recalibration on misaligned features spatially.\\
\indent
With aforementioned methods we proposed, we try our best to improve the performance of Chinese herbal recognition. Consequently, the contributions in this study can be concluded as follows:\\
1. We build and present the standard Chinese-Herbs recognition dataset (CNH-98), further, we build the corresponding tiny-Chinese-Herbs dataset (TCNH-98), which is used to train models for locally recognition of herbs.\\
2. We introduce both channel-wise and spatial attention mechanism into pyramid networks and further improve their structures to propose channel-wise competitive attention and spatial reference attention. The former focus on modeling channel dependencies between spatial and semantic information flows and the latter tends to recalibrate the misaligned features with spatial information flows for reference.\\
3. We first apply pyramid ConvNet to Chinese herbal recognition according to the characteristic of recognition task.\\
4. We conduct experiments on the datasets we proposed to validate the superior performance of presented models on the task of Chinese herbal recognition.
%------------------------------------------------------------------------
\section{Related Work}
\textbf{Feature Pyramid.} Feature pyramid network is proposed to get image features at different scales\cite{Lin2016Feature}, based on this motivation,
numerous methods with multi-level features in CNN have been proposed, such as RoI pooling\cite{Girshick2015Fast} or using skip-connection to construct pyramid\cite{Newell2016Stacked}. With RoI pooling on proposal region, HyperNet\cite{Kong2016HyperNet}, ParseNet\cite{Liu2015ParseNet} and ION\cite{Bell2016Inside} concatenate features of multiple layers before computing predictions and \cite{Chang2018Pyramid,Yang2018Learning} also aggregate context in different scales with spatial pooling. Feature pyramid like Stacked Hourglass network\cite{Newell2016Stacked} is the typical structure with skip-connection, which combines different levels features for key point estimation. Inspired by Hourglass Module, FPN\cite{Lin2016Feature} designs a network with strong semantic at all scales for object detection and FANet\cite{Zhang2017Feature} improves it further by augmenting lower-level feature maps. Several other approaches including PRM\cite{Yang2017Learning} for pose estimation, U-Net\cite{Ronneberger2017U} for segmentation and RON\cite{Kong2017RON} for object detection handling features at multi-level by skip connections. In our work, we introduce an attentional fusion method based on FPN\cite{Lin2016Feature} to competively model the relationship between spatial information and semantics for Chinese Herbal Recognition.
\\
\indent
\textbf{Attention in CNN.} With the trend of attention widely applied on the modeling process of CNNs\cite{Nguyen2018Attentive}, it is commonly used for two primary aspects: channel-wise attention \cite{Hu2017Squeeze} to explicitly model interdependencies between channels and the other one to re-weight the image spatial signals\cite{Wang2017Residual,Li2018Harmonious,Woo2018CBAM,Zheng2017Learning}. Furthermore, some models combine both spatial and channel-wise attention, such as SCA-CNN\cite{Chen2017SCA,Linsley2018Global}. However, the mentioned models are limited on local region. To solve this problem, self-attention \cite{Wang2017Non,Fu2018Dual} is proposed to capture long-range dependencies between local and global. Additionally, there are some attention models based on domain knowledge \cite{Chen2017Attentive,Choi2017GRAM}. Interaction-aware pyramid\cite{Du2018Interaction} also introduce attention to the network for modeling long-range relationship. Different from \cite{Du2018Interaction}, our proposed attention mechanism based on the specific structure of FPN\cite{Lin2016Feature} explicitly models a trade off between spatial and semantic informations for Chinese Herbal recognition.\\
\indent
\textbf{CNN Applied on Tasks like Herbal Recognition.} There are some similar tasks using CNN with Chinese Herbal recognition such as plants recognition \cite{toth2016deep}, which mainly focus on leaf recognition\cite{hu2018multiscale,Ayaz2017Leaf}. Moreover, another similar tasks like flower recognition\cite{Gogul2017Flower,Xia2017Inception} can also use CNN to achieve. As far as we konw, there has been no one using CNN to recognize Chinese Herbs so far and we propose this approach firstly.
%-------------------------------------------------------------------------
\section{Chinese-Herbs Dataset Collection}
\begin{figure*}[ht]
\centering
\includegraphics[scale=0.7]{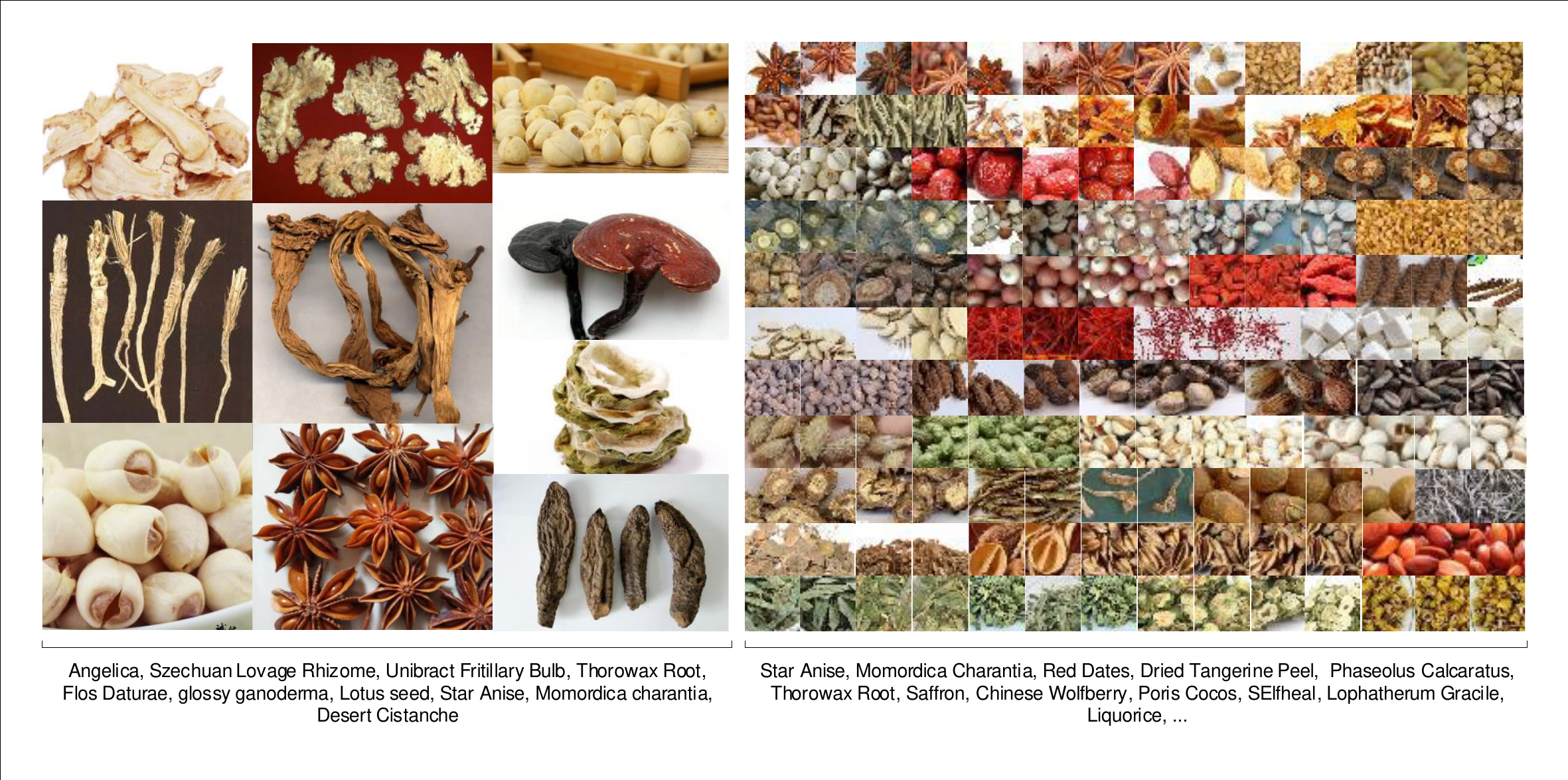}
\caption{Examples from the proposed CNH-98 dataset (left) and TCNH-98 dataset (right).}
\label{fig_3}
\end{figure*}
The \textit{Chinese-Herbs Dataset} (CNH-98) is a collection of 9184 images of 98 categories covering the common Chinese herbs. Furthermore, we make a crop of each image into serveral tiny images without overlapping to construct a \textit{Tiny-Chinese-Herbs Dataset} (TCNH-98) including 51198 images, because each image always contains multiple repeated herbs. These two datasets are divided randomly into training and validation sets with the proportion of 4:1. Fig. \ref{fig_3} shows some examples of CNH-98 (left) and their crop TCNH-98 (right). The sample datasets are available\footnote{\url{https://github.com/scut-aitcm/Chinese-Herbs-Dataset}}. \\
\subsection{Chinese-Herbs Dataset}
In this dataset, most of the images were acquired by taking photos ourselves in the medicinal herbs stores, hospitals and so on. And the others were collected from the Google images\cite{GoogleImage}. The smallest dimension of images is about 250 pixels. Each class contains 94 images on average and more than 41 classes include over 100 images. In order to ensure the availability and matching of labels and data, the labels were reviewed by the human annotators. \\
\subsection{Tiny-Chinese-Herbs Dataset}
Tiny dataset was sampled from above CNH-98 dataset with the size of $32\times 32$ and we ensured that there was no overlapping. Considering that there are some  factors interfering with the quality of the image, such as blank place in the origin and so on, we dropped some images in the following conditions, as judged by the annotators: (i) the images were blank or the proportion of herbs in images is too small, (ii) not contain herbs (like some containers or background), (iii) the annotators cannot recognize such as the parts of original herbs. Overall, we gained an average of 522 images per class and minimum 100 per class.

We need to make a statement that the Tiny-Chinese-Herbs dataset may bring more severe challenges in herbal recognition, due to the limited image size and incomplete features of herbs, although the scale of this dataset is bigger.
%-------------------------------------------------------------------------
\section{Competitive Attentional Fusion Pyramid Networks}
In this section, considering the characteristics of Chinese herbal recognition tasks, we first extend applications on \textit{Feature Pyramid Network} (FPN\cite{Lin2016Feature}) to Chinese herbal recognition tasks. Next, we propose a competitive attentional fusion mechanism based on the original FPN to adapt to the aforementioned tasks. Finally, in terms of existing problem of misaligned features, a spatial recalibtration method is proposed, which will be combined with the above attentional fusion mechanism.
\subsection{Apply FPN to Chinese Herbal Recognition}
\begin{figure}[ht]
\centering
\includegraphics[scale=0.35]{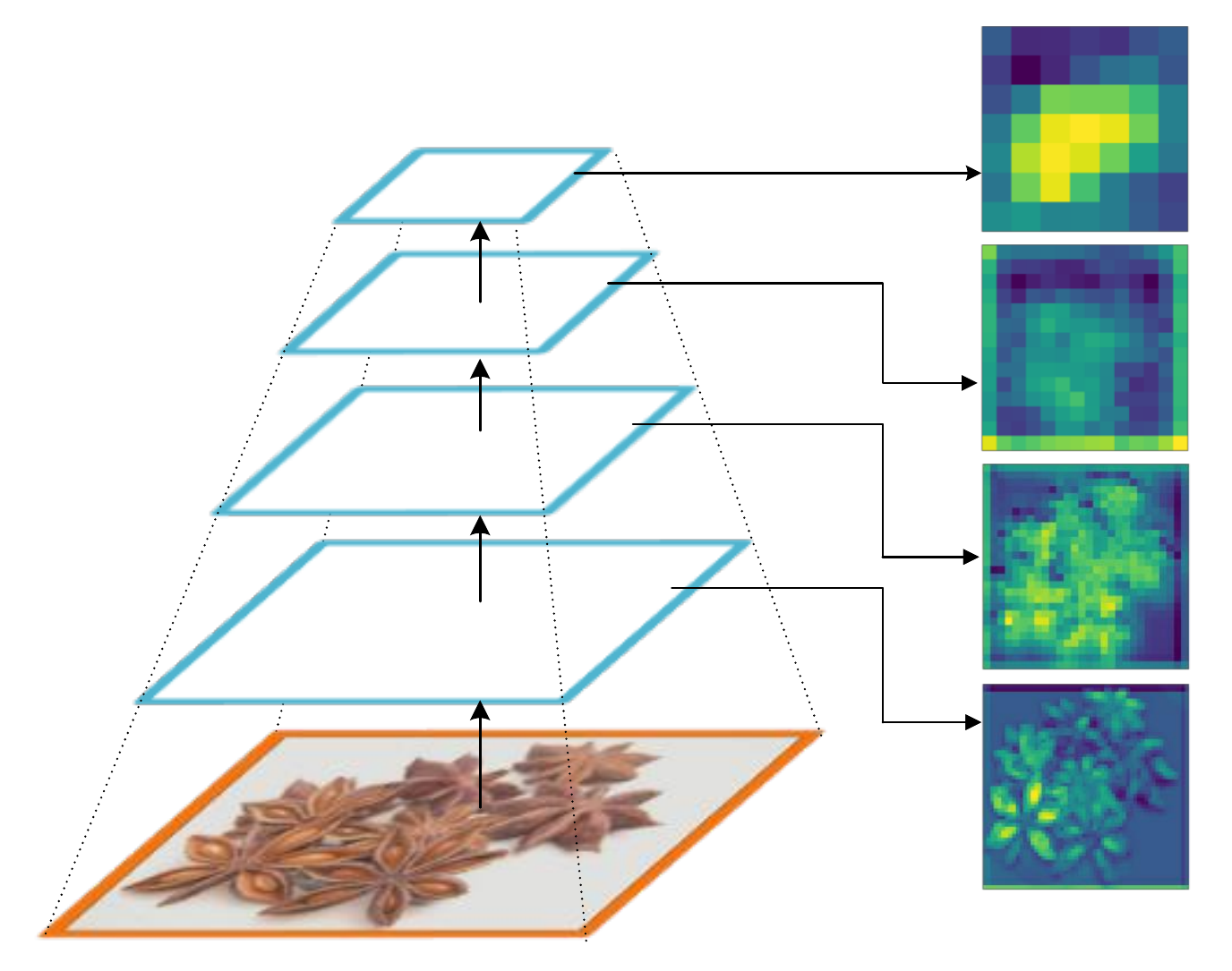}
\caption{FeaturePyramid. Here we extract feature maps from four levels of ResNet-18 to form a feature pyramid.}
\label{fig_4_1}
\end{figure}
For herbal recogniton tasks, there is a characteristic that the shapes of some herbs are so distinguishing that they are easy to be classified using the high-resolution features from the lower level of networks, while some herbs with similar shapes usually need to be classified by features from the higher level, which contain more fine-grained semantic informations. The feature maps from various layers of networks are shown in Fig. \ref{fig_4_1}. Therefore, we choose FPN\cite{Lin2016Feature} applied to Chinese herbal recognition, because FPN can fuse
multi-hierarchies features with its pyramid structure.\\
\indent Consisting of two pathway, a bottom-up pathway and a top-down pathway, and lateral connections, FPN can build a feature pyramid with high-level semantics throughout by naturally exploiting a ConvNet's pyramid feature hierarchy. The bottom-up pathway is the feed-forward computation of the backbone ConvNet, which results in a feature hierarchy containing feature maps at several scales with a scaling step of 2. And the top-down pathway generates higher resolution features by upsampling the last groups of feature maps by a factor of 2 on the bottom-up pathway. Here we record the output of upsampling as $X_{U_i} \in \mathbb{R}^{W_i \times H_i\times C_d}$. As opposed to the features on the same level from the bottom-up pathway, these feature maps are spaitally coarser, but semantically stronger, hence we natrually refer the bottom-up pathway to the spatial flow and the top-down pathway to the semantic flow. As described in the design of FPN, the output of last layer of $i^{th}$ level via lateral connections $X_{L_i} \in \mathbb{R}^{W_i \times H_i\times C_d}$ merges with the corresponding feature maps with the same size from top-down pathway, as follows:
\begin{equation}\label{equ:fpn_origin}
X_{P_i}=X_{L_i}+X_{U_i},
\end{equation}
where $X_{P_i}$ will be fed into the next upsampling process. The result is a fusion feature pyramid that has strong semantic and spatial informations at all scales.
%-------------------------------------------------------------------------
\subsection{Competitive Attention between Spatial and Semantic Flows}
\begin{figure*}[ht]
\centering
\includegraphics[scale=0.45]{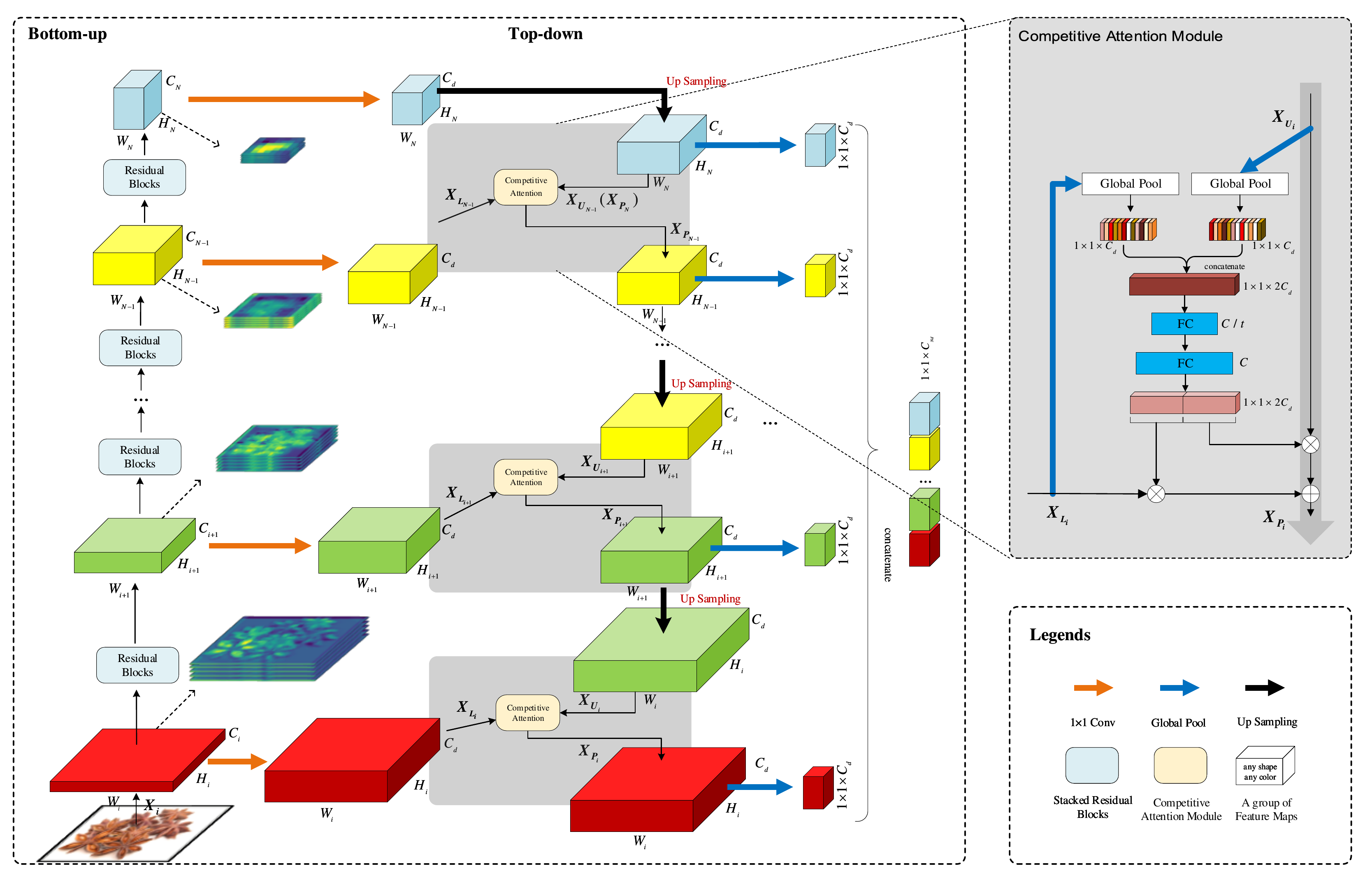}
\caption{Overview of Competitive Attentional Fusion Pyramid Network and Competitive Attention modules. }
\label{fig_4_2}
\end{figure*}
The aforementioned fusion mode of features in FPN is to indiscriminately treat spatial and semantic flows at all scales, which is likely to cause redundencies in fusion features. From an intuitional point of view, we propose a competive attention mechanism that allows the network to explicitly modeling the competition between spatial and semantic flows in the process of fusion, such that the network can selectively emphasis richer semantic or spatial features and suppress redundant ones. \\
\indent To achieve this, we gain the global information $\hat{\mathrm{u}}_{spa}$ and $\hat{\mathrm{u}}_{sem}$, embedding from feature maps via lateral connections of spatial flow $X_L=[x_L^1,x_L^2,\ldots,x_L^C]$ and upsampling feature maps of semantic flow $X_{U}=[x_U^1,x_U^2,\ldots,x_U^C]$ respectively:
\begin{equation}\label{equ:se_gap_spa}
\hat{\mathrm{u}}^c_{spa}=F_{sq}(\mathrm{x}_L^c)=\frac{1}{W \times H}\sum^W_{i=1}\sum^H_{j=1}x_L^c(i,j),
\end{equation}
\begin{equation}\label{equ:se_gap_sem}
\hat{\mathrm{u}}^c_{sem}=F_{sq}(\mathrm{x}_U^c)=\frac{1}{W \times H}\sum^W_{i=1}\sum^H_{j=1}x_U^c(i,j),
\end{equation}
where $F_{sq}(\cdot)$ denotes the operation of global pooling. The combination of $\hat{\mathrm{u}}_{spa}$ and $\hat{\mathrm{u}}_{sem}$ will be used as joint input for the excitation operation to capture channel-wise dependencies between spatial and semantic flows:
\begin{equation}\label{equ:se_ex}
\begin{split}
\mathrm{s}&=F_{ex}([\hat{\mathrm{u}}^1_{spa},\hat{\mathrm{u}}^2_{spa},\ldots,\\
& \qquad \qquad \hat{\mathrm{u}}^C_{spa} ,\hat{\mathrm{u}}^1_{sem}, \hat{\mathrm{u}}^2_{sem},\ldots,\hat{\mathrm{u}}^C_{sem}],\mathrm{w}_{ex})\\
&=F_{ex}([\hat{\mathrm{u}}_{spa},\hat{\mathrm{u}}_{sem}],\mathrm{w}_{ex})\\
&=\sigma (ReLU([\hat{\mathrm{u}}_{spa},\hat{\mathrm{u}}_{sem}],\mathrm{w}_1),\mathrm{w}_2),
\end{split}
\end{equation}
where $[\cdot]$ refers to the concatenation of the feature-maps produced in the above squeeze operation from two flows, and parameters $\mathrm{w}_1\in \mathbb{R}^{\frac{2C}{t} \times 2C}$ and $\mathrm{w}_2 \in \mathbb{R}^{2C \times \frac{2C}{t}}$. The result $s$ of \textit{Excitation} operator $F_{ex}(\cdot)$ is $[\mathrm{s}_1,\mathrm{s}_2,\ldots,\mathrm{s}_C,\mathrm{s}_{C+1},\mathrm{s}_{C+2},\ldots,\mathrm{s}_{2C}]$ that will divide into two parts to rescaling the weights of features $X_L$ and $X_U$ respectively as follows:
\begin{equation}\label{equ:se_scale_spa}
\begin{split}
\tilde{\mathrm{x}}^c_{spa} &=F_{scale}(\mathrm{s}^c_{spa},\mathrm{x}_L^c)\\
&=F_{se}(X_L)[\cdot]\times X_L=\mathrm{s}^c_{spa}\cdot \mathrm{x}_L^c,
\end{split}
\end{equation}
\begin{equation}\label{equ:se_scale_sem}
\begin{split}
\tilde{\mathrm{x}}^c_{sem} &=F_{scale}(\mathrm{s}^c_{sem},\mathrm{x}_U^c)\\
&=F_{se}(X_U)[\cdot]\times X_U=\mathrm{s}^c_{sem}\cdot \mathrm{x}_U^c,
\end{split}
\end{equation}
where $\mathrm{s}_{spa}$ refers to $[\mathrm{s}_1,\mathrm{s}_2,\ldots,\mathrm{s}_C]$ and $\mathrm{s}_{sem}$ means $[\mathrm{s}_{C+1},\mathrm{s}_{C+2},\ldots,\mathrm{s}_{2C}]$.
\indent The competition between spatial and semantic flows is modeled by the \textit{Competitive Attention} module proposed above and react to each channel of both spatial and semantic flow. On the one hand, the aforementioned mergence mode of features can be regarded as a adaptive competition between two flows and its recalibration depends on two flows adaptively to dynamically adjust the complement weights for each other. On the other hand, a trade off between spatial and semantic flows is indicated. Finally, the Competitive Attention module is reformulated as:
\begin{equation}\label{equ:se_scale}
X_P=F_{se}(X_L)\cdot X_L+ F_{se}(X_U)\cdot X_U,
\end{equation}
\indent
Fig. \ref{fig_4_2} shows the overview of our Competitive Attentional Fusion Pyramid Network and its more details in Competitive Attention module. It is concluded that the difference between the typical SE and the Competitive Attention is that based on the particular structure and meanings of FPN we simultaneously introduced two flows into SE to model their channel relationship competitively and trade off, and meanwhile we adjust two flows at the same time.
%-------------------------------------------------------------------------
\subsection{Spatial Reference Recalibration}
\begin{figure}[ht]
\centering
\includegraphics[scale=0.4]{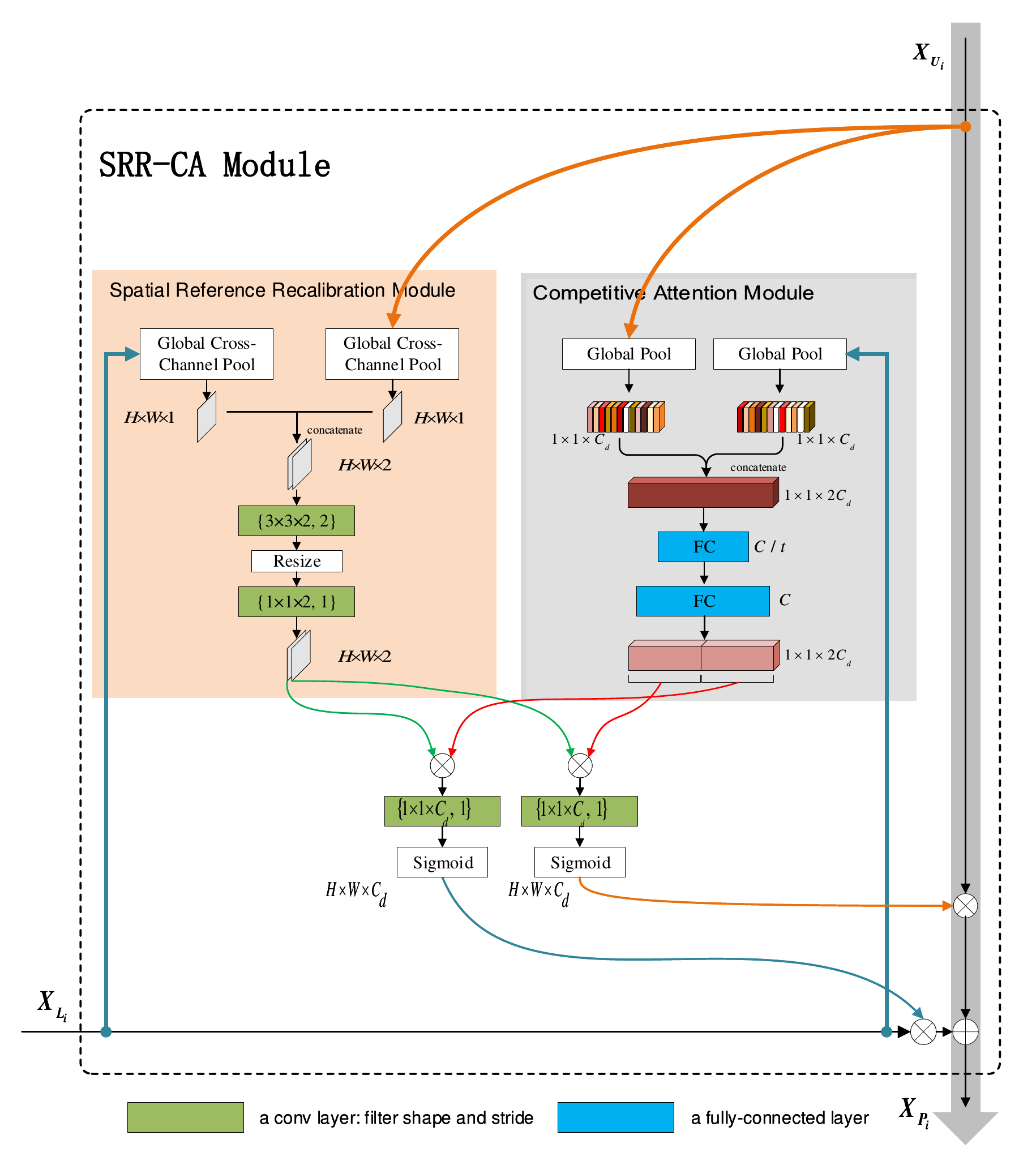}
\caption{Spatial Reference Recalibration module and its combination with Competitive Attention module. The Batch Normalisation\cite{ioffe2015batch} (BN) (attached to two conv layers after tensor multiplication) is not shown for brevity. We resize feature maps in SRR module with the factor of 2 by bilinear interpolation.}
\label{fig_4_3}
\end{figure}
As we discuss above, the upsampling features before merging are spatially coarser because they are products of several downsampling or upsampling operators. In other words, their spatial informations such as location are less accurate and even misaligned. That is also why we need fusion features. \\
\indent However, it should be noted that the above amalgamation means of element-wise addition extremely rely on the spatial informations, thus it is likely that the fusion features merged in this way are sub-optimal. Consequently, we introduced a method to spatially recalibrate the misaligned features through modeling spatial attention on pixel-level with spatial flows for reference. As discussed in \textit{Harmonious-Spatial-Attention}\cite{Li2018Harmonious} (HA), similarly we compresses the feature maps in the following ways (global cross-channel averaging pooling) to reduce parameters for the subsequent conv layer, but unlike HA we Simultaneously model two flows:
\begin{equation}\label{equ:ha_gcap_spa}
S^i_{spa}=\frac{1}{c}\sum^C_{j=1}{X_L}^i_{1:h,1:w,j},
\end{equation}
\begin{equation}\label{equ:ha_gcap_sem}
S^i_{sem}=\frac{1}{c}\sum^C_{j=1}{X_U}^i_{1:h,1:w,j},
\end{equation}
where $S^i_{spa}$ and $S^i_{sem}$ will be concatenated to $S^i=[S^i_{spa},S^i_{sem}]$ fed into the next conv layer of $3 \times 3$ filters with stride 2 and then resized to original size by bilinear interpolation with the factor of 2. Finally, we add the scaling conv layer of $1\times 1$ filters for reducing aliasing effect of bilinear upsampling. As a result, we gain 2 feature maps to rescale values of features from two flows on pixel-level respectively. In addition, this mechanism also contributes to the robustness of network which allows it to use different upsampling methods on the top-down pathway.\\
\indent Aming to combine the competitive attention with spatial recalibration, we further attach two $1\times 1\times C_d$ convolution layer after tensor multiplication on two flows respectively, since the two procosses are not mutually independent. Finally, we deploy the sigmoid operations to normalise. More details of SRR-module and its combination with Competitive attention are shown in Fig. \ref{fig_4_3}.
%-------------------------------------------------------------------------
\section{Experiments}

%-------------------------------------------------------------------------
\subsection{Implementation Details}
For fair comparison, each plain FPN and its corresponding CA, SRR and SRR-CA counterparts are trained with identical optimisation schemes. For CNH-98 and TCNH-98 datasets, we train our all models with three degrees of data augmentation: no data augmentation, standard data augmentation (+) and $mixup$\cite{Zhang2018mixup}, an advanced data augmentation technology. On CNH-98, the standard data augmentation (translation/mirroring) is adopted for training set and the 224x224 crop is randomly sampled. All images normalized with mean values and standard deviations. When testing, our implementation follows the practice in \cite{Hu2017Squeeze}. On TCNH-98, we follows the standard practice and data augmentation in \cite{he2016deep} for CIFAR. All models were trained by optimizer SGD with 0.9 Nesterov momentum from scratch. \\
\indent During training on CNH-98, we train our models with  batch size 64 and 300 epochs for standard augmentation and mixup, 120 epochs for no augmentation. The learning rate is initialized to 0.1 and divided by 5 at epochs 120, 200, 260 for standard augmentation and $mixup$ and at epochs 30, 60, 90 for no augmentation , and weight decay are adopted with 0.0005 and 0.0001 respectively. In particular, we train models for $mixup$ on the last 20 epochs with traditional strategy.\\
\indent During training on TCNH-98, our models are trained for 300 epochs with batch size 128 and the initial learning rate is 0.1 and is divided by 10 at 100th, 150th, 200th epochs. We also set the weight decay as 0.0001 following \cite{he2016deep} for CIFAR. Especially, learning rate during training without data augmentation was divided by 5 at epochs
30, 60, 90.
%-------------------------------------------------------------------------
\subsection{Results of Chinese Herbal Recognition}

\begin{table*}[ht]
\center
\begin{tabular}{l c c c c c }
\multicolumn{6}{l}{\textbf{$\boldsymbol{D_1}$: Chinese-Herbs}}\\
\hline
\multirow{1}{5.5cm}{Model}
~&backbone depth&parames&CNH-98&CNH-98+&CNH-98 $mixup$\\
\hline
pre-act ResNet-18\cite{he2016identity}&18&11.7M&74.5&91.7&93.3\\
FPN-pre-act ResNet-18\cite{Lin2016Feature}&18&13.3M&74.7&91.9&93.5\\
\hline
  FPN-CA-18(Ours) &18&13.4M&75.3&92.9&\textbf{94.2}\\

  FPN-SRR-18(Ours)&18&13.3M&72.5&92.5&93.8\\

  FPN-SRR-CA-18(Ours)&18&13.8M&\textbf{76.8}&\textbf{93.5}&94.1\\

  \hline
  FPN-pre-act ResNet-34\cite{Lin2016Feature}&34&23.4M&75.1&92.3&94.1\\
  \hline
  FPN-CA-34(Ours) &34&23.5M&76.1&93.5&94.6\\

  FPN-SRR-34(Ours)&34&23.4M&-&92.7&-\\

  FPN-SRR-CA-34(Ours)&34&23.9M&\textbf{76.3}&\textbf{93.8}&\textbf{94.8}\\
  \hline
\multicolumn{6}{l}{\textbf{$\boldsymbol{D_2}$: Tiny-Chinese-Herbs}}\\
\hline
\multirow{1}{5.5cm}{Model}
~&backbone depth&parames&TCNH-98&TCNH-98+&TCNH-98 $mixup$\\
\hline
pre-act ResNet-20\cite{he2016identity}&20&0.28M&63.0&74.8&72.8\\
FPN-pre-act ResNet-20\cite{Lin2016Feature}&20&0.31M&63.1&75.2&72.9\\
\hline
  FPN-CA-20(Ours) &20&0.31M&62.8&75.8&73.6\\

  FPN-SRR-20(Ours)&20&0.31M&-&75.5&73.3\\

  FPN-SRR-CA-20(Ours)&20&0.31M&\textbf{63.8}&\textbf{75.8}&\textbf{73.7}\\

  \hline
  FPN-pre-act ResNet-56\cite{Lin2016Feature}&56&0.89M&\textbf{64.3}&77.4&77.4\\
  \hline
  FPN-CA-56(Ours) &56&0.89M&63.1&\textbf{77.7}&76.7\\

  FPN-SRR-CA-56(Ours)&56&0.90M&62.8&77.6&\textbf{77.6}\\
  \hline

\end{tabular}
\caption{Accuracy rates(\%) of different methods on datasets CNH-98 and TCNH-98, the best records of our models are \textbf{bold}. We compare our models with the original FPN and its backbone networks, trained with either no data augmentation, standard augmentation (+) and $mixup$.}
\label{tab5_2}
\end{table*}
We evaluate our methods on the CNH-98 and TCNH-98 datasets with pre-act ResNet\cite{he2016identity} for backbone networks and the results of contrastive experiments for FPN with/without CA and SRR-CA modules are shown in Table. \ref{tab5_2}, and we can make a summary as follows:\\
\indent
First of all, as shown in $D_1$ and $D_2$ in the Table. \ref{tab5_2}, we can see FPN indeed gets a better results than pre-act ResNet whether on CNH-98 or TCNH-98, which verifies the guess in Section 4.1 that FPN is more suitable to accomplish the task of Chinese herbal recognition, and here we record the experiment on FPN as baseline. Furthermore, for both CNH-98 and TCNH-98, FPN-CA can achieve superior performance than baseline and FPN-SRR-CA can further improve performance across different depth or keep the effect at least without too much extra parameters.\\ \indent
Secondly, FPN-SRR almost can exceed FPN except on CNH-98 without data augmentation, proving the effectiveness of SRR modules in most case and suggesting that CA and SRR modules are not two separate processes but need to model jointly, hence it is reasonable to attach $1\times 1$ convolution layer after combination of SRR and CA modules. For the reason of performance of SRR on CNH-98 with no augmentation, we infer that there is an overfitting phenomenon owing to the small size of CNH-98 dataset. Moreover, on CNH-98 dataset, compared with FPN-34, FPN-SRR-CA-18 even increases validation accuracy rates by 1.7\% for no augmentation, 1.2\% for standard augmentation and achieve or slightly go beyond of FPN-34 for mixup. In particular, FPN-SRR-CA-18 has higher accuracy rates than FPN-SRR-CA-34 for no augmentation, for which we infer that the depth 34 of networks for small dataset like CNH-98 is too deep to fit and our CA and SRR-CA modules can reduce overfitting as well as improving the generalization ability of models thus perform better with deeper networks. On the contrary, during training on TCNH-98 that consists of 40958 images with standard augmentation and $mixup$, we notice that there is an underfitting for the depth 20 of networks, which indicates the representation of the models with depth 20 is too limited, and we increased the depth of networks, which can reduce this phenomenon, proving the performance of models with deeper networks can get better.\\
\indent
The $mixup$\cite{Zhang2018mixup} can be seen as an advanced method of data augmentation. However, for TCNH-98 dataset, models with $mixup$ achieve the worse results, for which we argue that $mixup$ as augmentation approaches would further aggravate underfitting, leading to a worse result natrually. Due to the limited representation of networks with depth 20, actually TCNH-98 dataset is suitable for deeper networks, proved by results of experiments on models with depth 56, which reduces underfitting.
%-------------------------------------------------------------------------
\subsection{Further Analysis and Discussion}
\begin{figure*}[ht]
\centering
\includegraphics[scale=0.6]{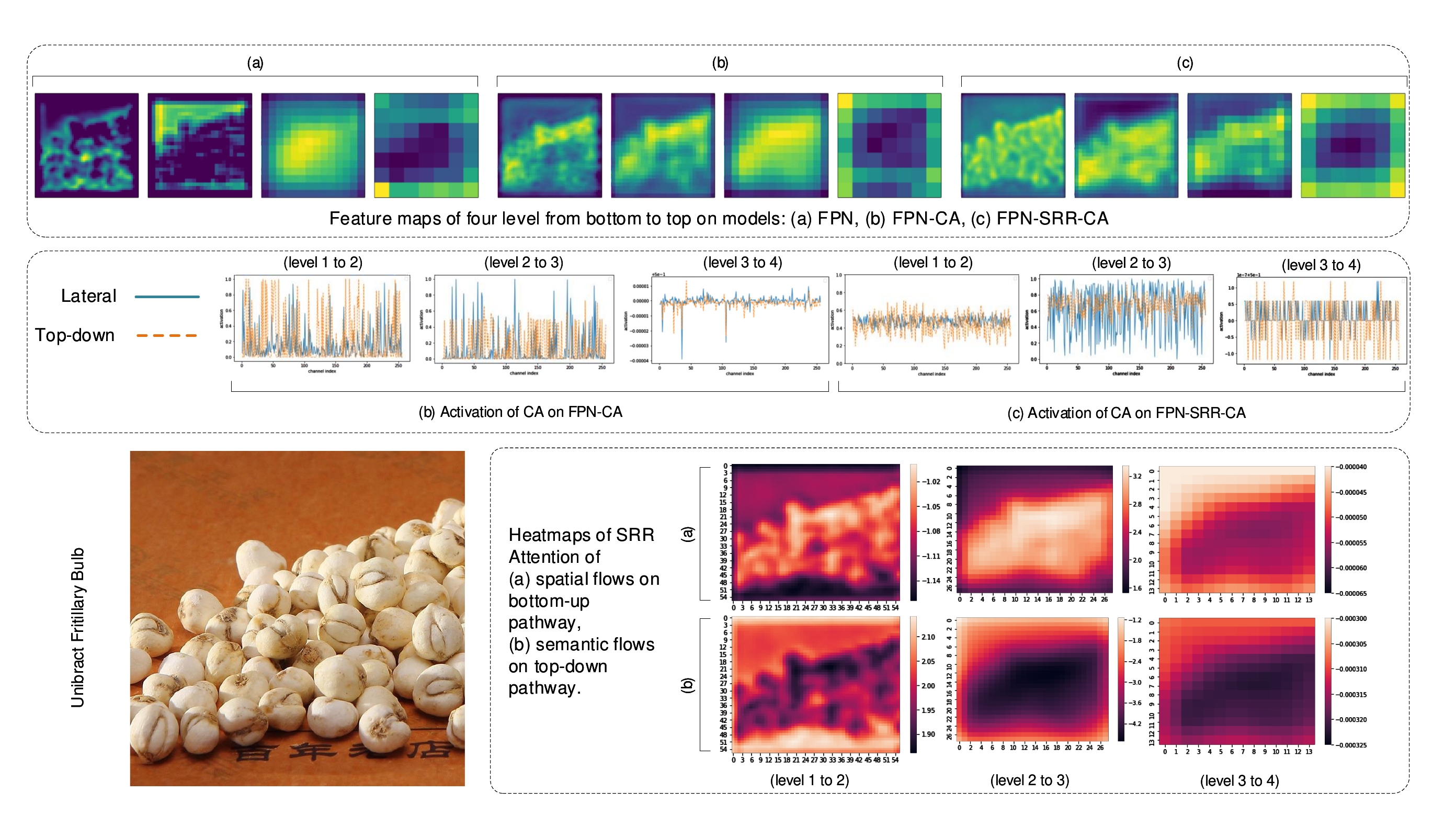}
\caption{Top: internal feature maps of an example from four levels on three models: (a) FPN, (b) FPN-CA, (c) FPN-SRR-CA. Middle: the activation values (solid lines for bottom-up pathway via lateral connections and dotted lines for top-down pathway) of competitive attention on (b) and (c). Bottom: the heatmaps of SRR attention.}
\label{fig_5_3}
\end{figure*}
The analysis of last section 5.2 has proven the effectiveness of CA and SRR-CA modules. In this section, from an intuitive angle of view, we discuss the effects of our approaches. The internal feature maps from different levels of three models, FPN-18, FPN-CA-18, FPN-SRR-CA-18, are shown in the top part of Fig. \ref{fig_5_3}, from which we can conclude that our methods can strengthen the representation of networks. By observing the representation of feature maps, the previous layers of FPN almost extract contour features, while the features are increased with more detailed informations using our FPN-CA models, compared with feature maps of FPN with/without CA on level 1 and 2 in Fig. \ref{fig_5_3}. It is worth mentioning that the features extracted by the models with CA modules are more sparse and accurate, compared to the original FPN, especially for feature maps of level 3. Moreover, SRR-CA modules can further spatially recalibrate the misaligned feature maps, mainly for deeper features, typically shown in level 3 of Fig. \ref{fig_5_3}, which makes the features with stronger spatial informations and richer in semantic. Additionally, we statistics the distributions of the activation of CA modules on FPN-CA and FPN-SRR-CA models, and we can see that the attentional activation values of CA and SRR modules are very vigorous and distinguish, and the heatmap of SRR modules can reconstruct the distribution of the origin, which suggests that our methods indeed contribute to re-weighting and recalibrating features.\\
\indent
As shown in the distribution of channel-wise attentional outputs, we can see the activation values of features from deeper layers are always uniform and tend to 0.5, for the reason that features from deeper layers have been adjusted during training, thus CA modules perform less adjustment. It is noticed that the activation values on the deepest level of spatial flow are almost higher than the ones from semantic flow, while from deep to previous, the activation from semantic flow would stand out from the competition gradually. This confirmes our conjecture that high-level features is spatially coarser and strongly semantic, in contrast to low-level features, and simultaneously indicates the mechanism we proposed can complement spatial or semantic informations for requirements of different levels. Correspondingly, there are same conclusion on the analysi of heatmap activation of SRR modules. Compared with channel-wise attentional outputs between FPN-CA and FPN-SRR-CA, there is a trend that channel-wise activation of FPN-SRR-CA would be more stable than FPN-CA owing to the effectiveness of SRR, which enables features more accurate and the effects of SRR can be passed through the network.
%------------------------------------------------------------------------
\section{Conclusion}
In this paper, we firstly propose the standard Chinese Herbs dataset for recognition. Based on the characteristic of Chinese herbal recognition task, we introduce attention mechanism into pyramid networks to model channel relationship of features from various levels. Furthermore, we also improve channel-wise and spatial attention and propose competitive attention and spatial reference recalibration module, which respectively model channel dependencies between spatial and semantic flows in the process of feature fusion and spatially recalibrate the misaligned feature maps with spatial flow for reference. With improved pyramid network, we apply it to the Chinese herbal recognition and evaluate our methods on CNH-98 and TCNH-98 dataset we proposed as well as getting superior performance to the traditional pyramid networks.

{\small
\bibliographystyle{ieee}
\bibliography{egbib}
}
\clearpage

\appendixpage
\appendices
\section{Details of Chinese Herbs Datasets}
\subsection{Distributions of Examples in CNH-98 Dataset}
\begin{table}[!hbp]
\center
\begin{tabular}{l | c}
\hline
Main Categories&\multirow{1}{*}{Herbs Examples}\\
\hline
\multirow{3}*{Fruits \& Seeds} & Star Anise, Siraitia Grosvenorii,\\
~&Ginkgo, Chinese Wolfberry,\\ ~&SElfheal, Fructus Arctii, etc.\\
\hline
\multirow{3}*{Rhizome} & {Liquorice, Thorowax Root,}\\
~&Rhizoma Alismatis, \\
~&Unibract Fritillary Bulb, etc.\\
\hline
\multirow{3}*{Flowers} &{Saffron, Flos Daturae,}\\
~&{Cloves, Magnolia, Coltsfoot,}\\
~&{Flos Jasmine, Lily, etc.}\\
\hline
\multirow{2}*{Bark} &{Cinnamon, Cortex Moutan,}\\
~&{Eucommia Ulmoides, etc.}\\
\hline
\multirow{2}*{Thallophyte} &{Glossy Ganoderma, Tremella
,}\\
~&{Cordyceps Sinensis, etc.}\\
\hline
\multirow{2}*{Whole Herbs} &{Abrus cantoniensis,}\\
~&{Anoectochilus roxburghii, etc.}\\
\hline
Leaves&{Lophatherum Gracile, etc.}\\
\hline
Resin&{Frankincense, Myrrh, etc.}\\
\hline

%\multicolumn{6}{l}{}

\end{tabular}
\caption{Main categories of CNH dataset and their corresponding examples.}
\label{tab_herbs_details}
\end{table}

Chinese Herbs are usually acquired from natural plants and the parts of fungus and algae, and our Chinese-Herbs Dataset (CNH-98) is a collection of 9184 images of 98 classes, which can be divided into 8 categories including Fruits \& Seeds, Rhizome, Flowers, Bark, Thallphyte, Whole Herbs, Leaves, Resin, whose examples are shown in Table. \ref{tab_herbs_details} correspondingly.\\
\begin{figure*}[ht]
\centering
\includegraphics[scale=0.65]{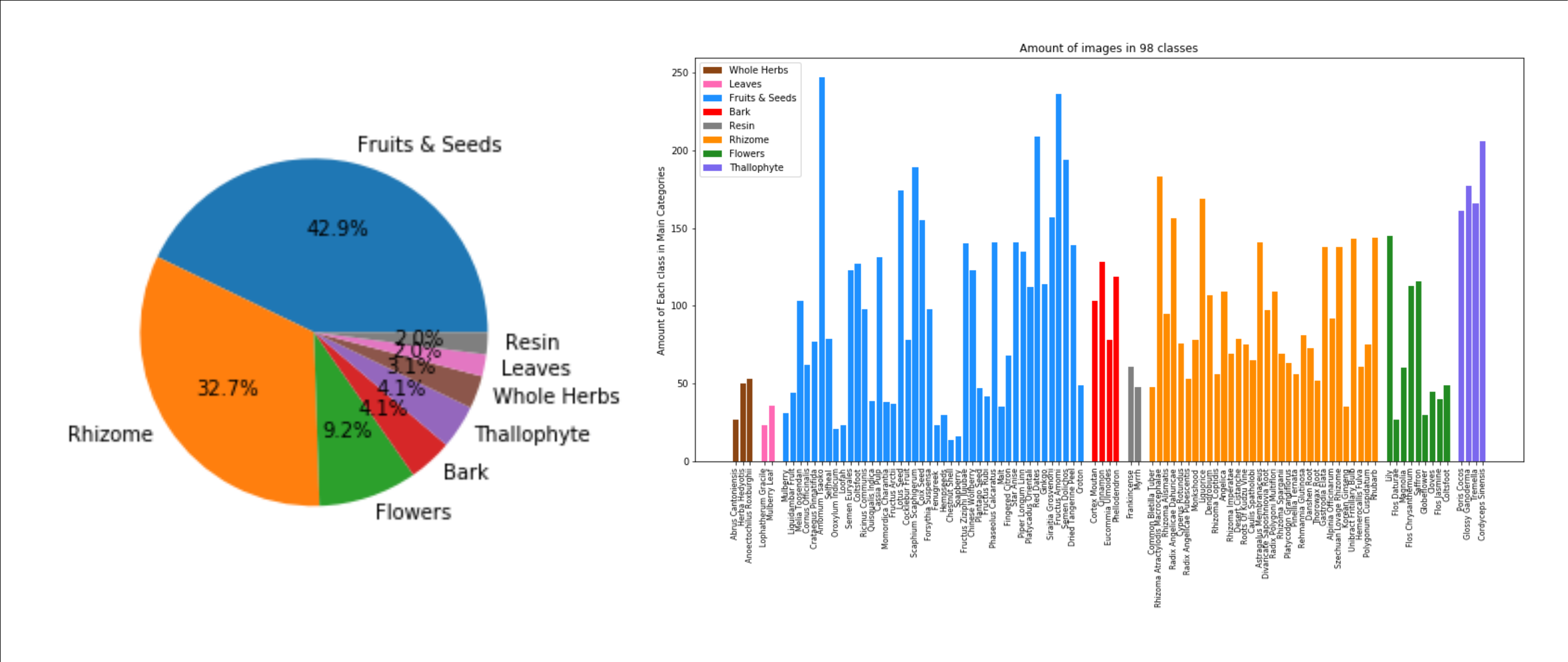}
\caption{Distribution of Chinese Herbs Categories (left) and amount of images for each classes in CNH-98 (right).}
\label{fig_pie_bar}
\end{figure*}
\indent
Fig. \ref{fig_pie_bar} (left) has shown the distibution of number of Chinese herbs classes in the 8 categories, where a majority of classes are Fruits \& Seeds and Rhizome, including 42 and 32 classes respectively. It can be seen that the CNH-98 dataset is relatively unbalanced. Moreover, as shown in Fig. \ref{fig_pie_bar} (right) , there is an unbalance of images quantities between 98 classes, the largest number of which is 247 images of Amomum Tsaoko and the least is 14 images of Chestnut Shell in Fruits \& Seeds.

\subsection{Exhibition of Main Categories}
\begin{figure*}[ht]
%\centering
\includegraphics[scale=0.83]{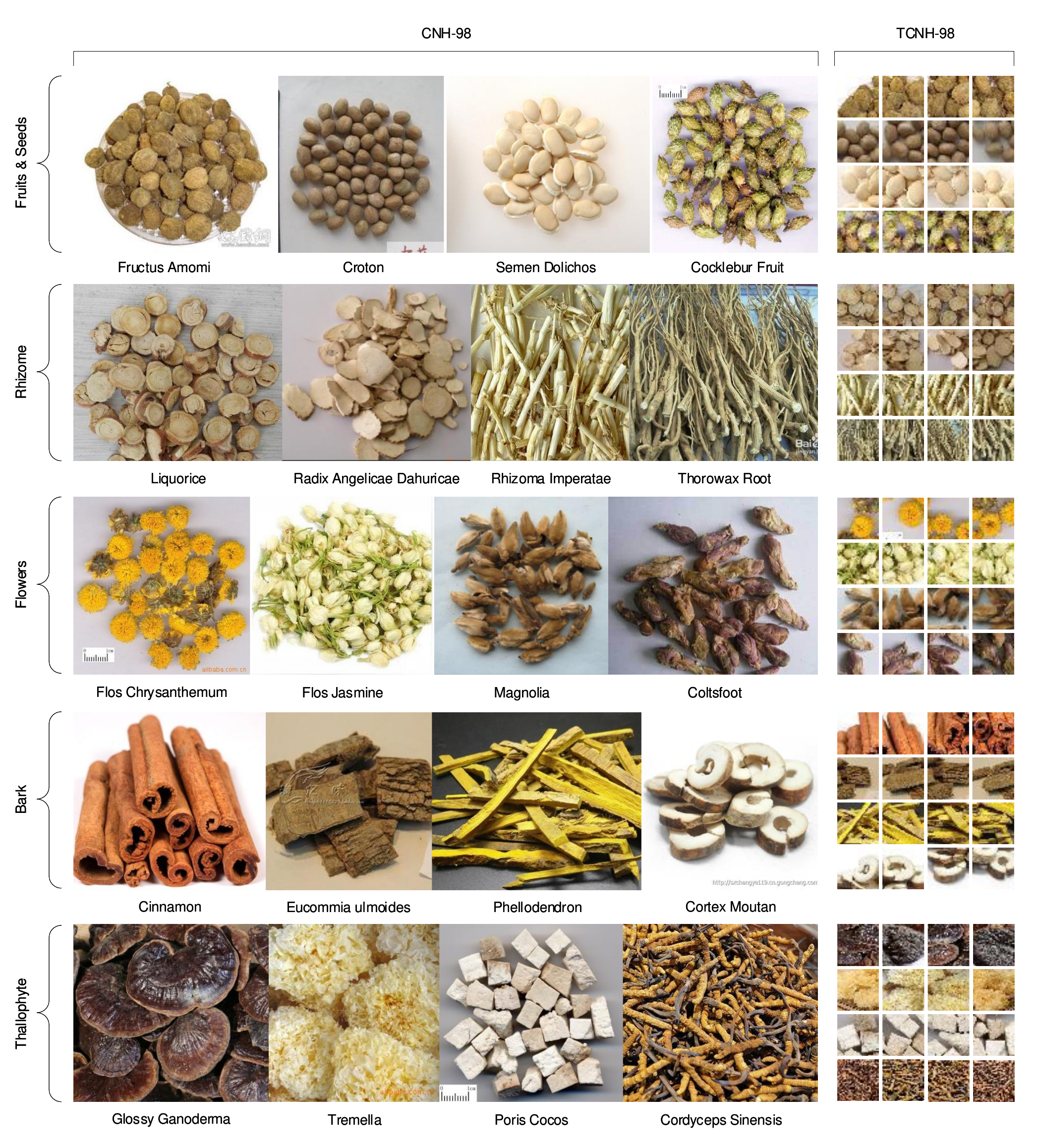}
\caption{Examples of Main Categories in CNH-98 and their Corresponding Examples in TCNH-98. From left to right, the left examples in CNH-98 corresponds to the right in TCNH-98 from top to bottom.}
\label{fig_exhibition}
\end{figure*}
In this section, we exhibit the examples of primary categories in CNH-98 and their corresponding cropping examples in TCNH-98, as shown in Fig. \ref{fig_exhibition}. From the exhibition in Fig. \ref{fig_exhibition}, we can see that although examples in TCNH-98 are just local parts, each example in TCNH-98 almost contains one herb with integrated shape at least, thanks to repeatability of examples in CNH-98. Furthermore, the shapes of Chinese herbs in various categories are extremely distinguishing, while the appearances of various classes in the same categories are similar, which is just the motivation of our proposed methods that the herbs with distinguishing shape can be classified by features from earlier layers of network, while the herbs with similar shape but different details need to be recognized by more semantic features from deeper levels.
\section{Evaluate with Other Upsampling Strategies}
\begin{table*}[ht]
\center
\begin{tabular}{l c c c c}
\hline
\multirow{1}{7cm}{Model} & Backbone Depth & Nearest & Deconvolution(\# params) & Bilinear*\\
\hline
FPN-18 & 18 & 91.0 & 90.9(16.5M) & 91.9\\

FPN-34 & 34 & 91.5 & - & 92.3\\
\hline
FPN-SRR-CA-18 (ours) & 18 & 93.4 & 93.0(16.9M) & 93.5\\
FPN-SRR-CA-34 (ours) & 34 & 93.6 & - & 93.8\\
\hline
FPN-SRR-18 (ours) & 18 & 92.3 & 92.3(16.5M) & 92.5\\
FPN-SRR-34 (ours) & 34 & 92.7 & - & 92.7\\
\hline
\end{tabular}
\caption{\textbf{Accuracy rates(\%)}. Compare our models(*) with FPN using different upsampling strategies.}
\label{tab_upsample}
\end{table*}
In order to validate the robustness of our models for different upsampling strategies, which mentioned in Section 4.3, we evaluate our methods with other upsampling methods including nearest neighbor and deconvolution and the results are shown in Table. \ref{tab_upsample}. By analyzing the results, we can conclude that the accuracy rates of FPN fluctuate more greatly than our models and the maximum discrepancy is by 1\% while only 0.1~0.5\% for our models. Even only using spatial attention SRR modules, our models can perform more stable, which reflects our SRR modules contribute to adapt for various upsamping strategies and perform more robustly.
\section{Further Analysis of Intermediate Results in FPN-CA/SRR-CA}
\begin{figure*}[t]
\centering
\includegraphics[scale=0.8]{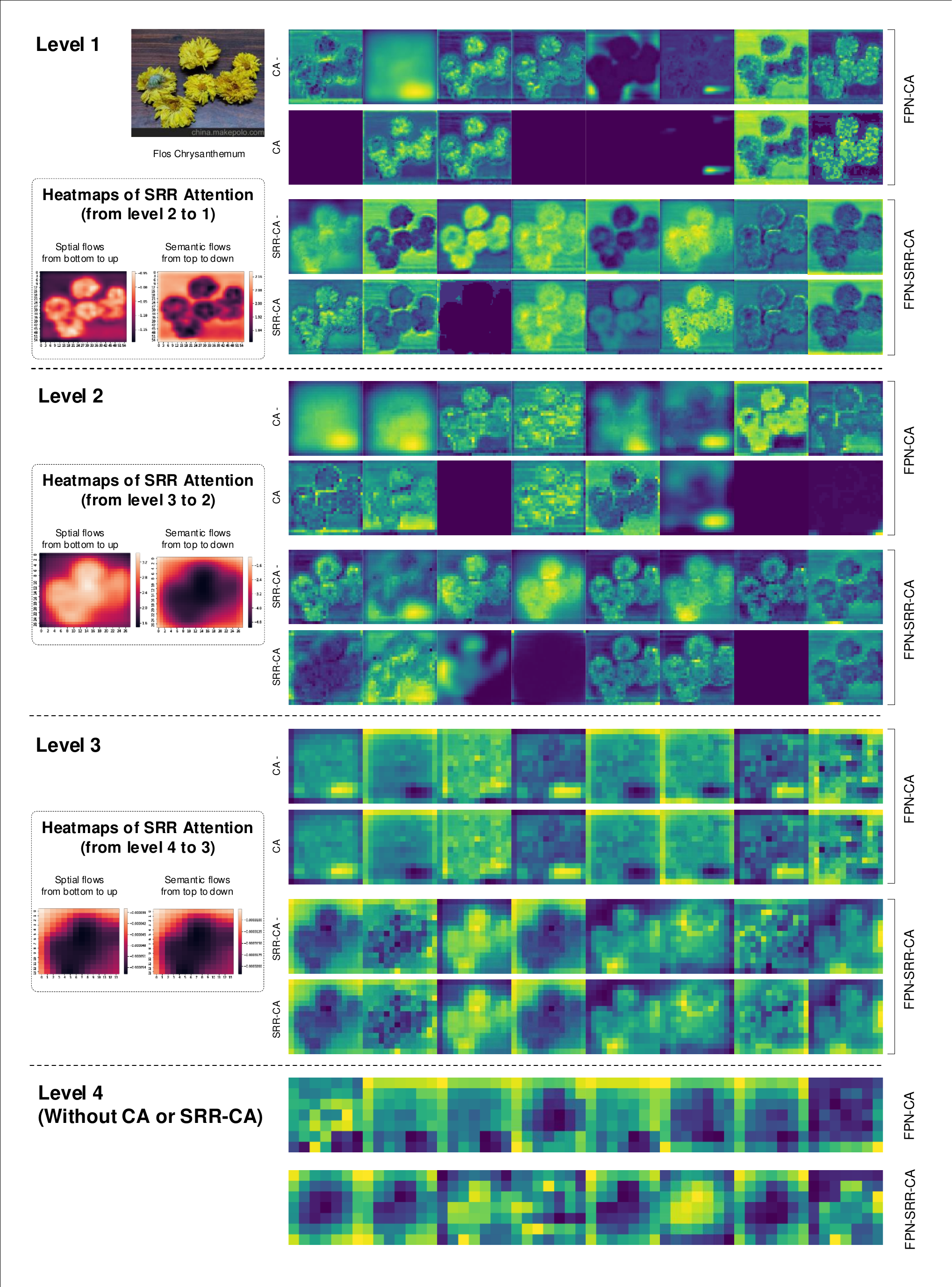}
\caption{\textbf{Flos Chrysanthemum: } intermediate features on various channels from 4 levels of models FPN-CA/SRR-CA-18, and their heatmaps of SRR Attention. Features on level 4 are initial on the top-down pathway, thus they are not fusion features. For each block (excepted for level 4), the \textbf{top (-)} are not re-scaled by attention and the \textbf{bottom} refers features reweighted on the corresponding channel.}
\label{fig_feature6}
\end{figure*}
\begin{figure*}[t]
\centering
\includegraphics[scale=0.8]{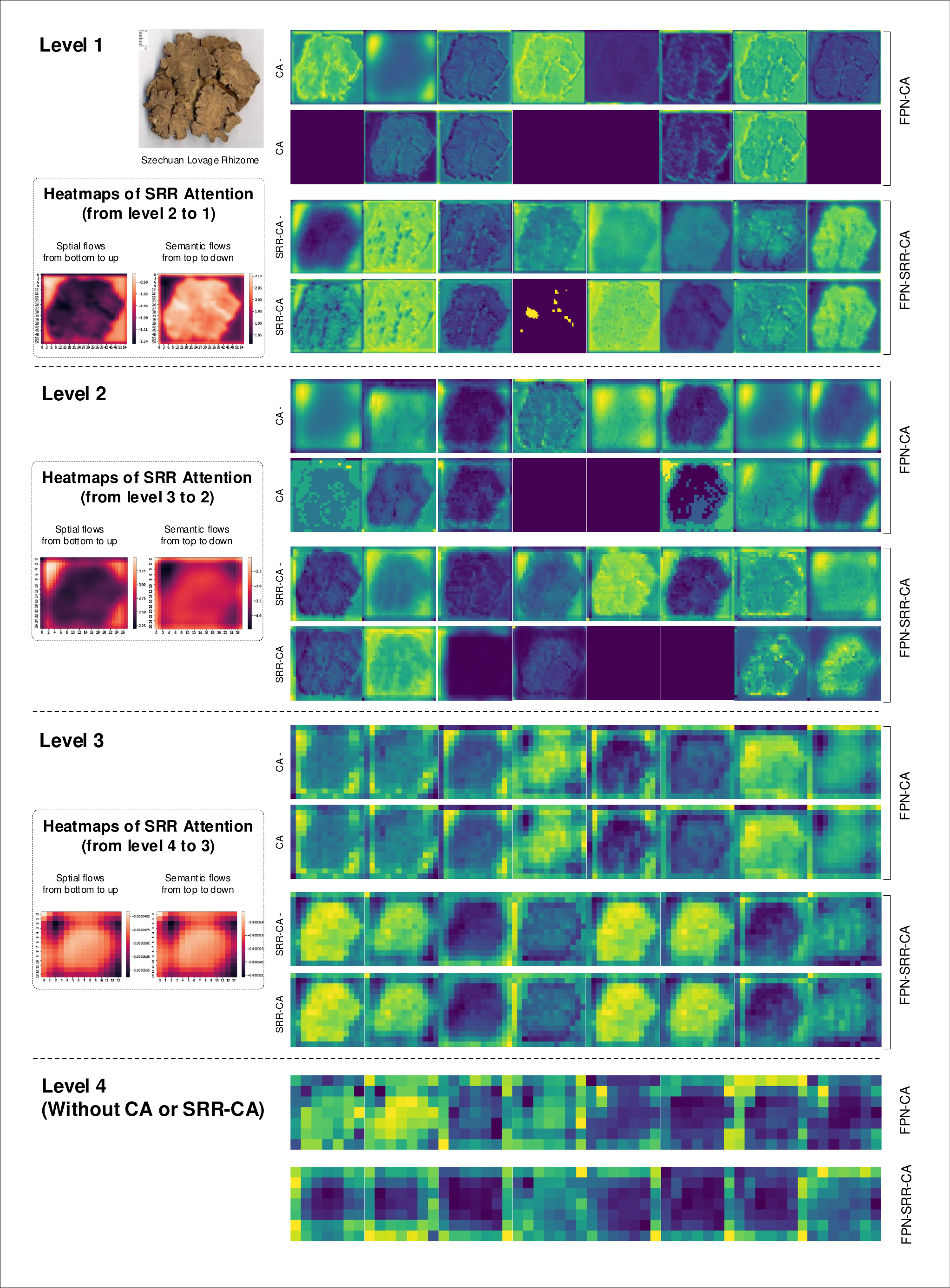}
\caption{\textbf{Szechuan Lovage Rhizome: }intermediate features on various channels from 4 levels and their heatmaps of SRR Attention. For each block (excepted for level 4), the \textbf{top (-)} are not re-scaled by attention and the \textbf{bottom} refers features reweighted on the corresponding channel.}
\label{fig_feature23}
\end{figure*}
\begin{figure*}[t]
\centering
\includegraphics[scale=0.8]{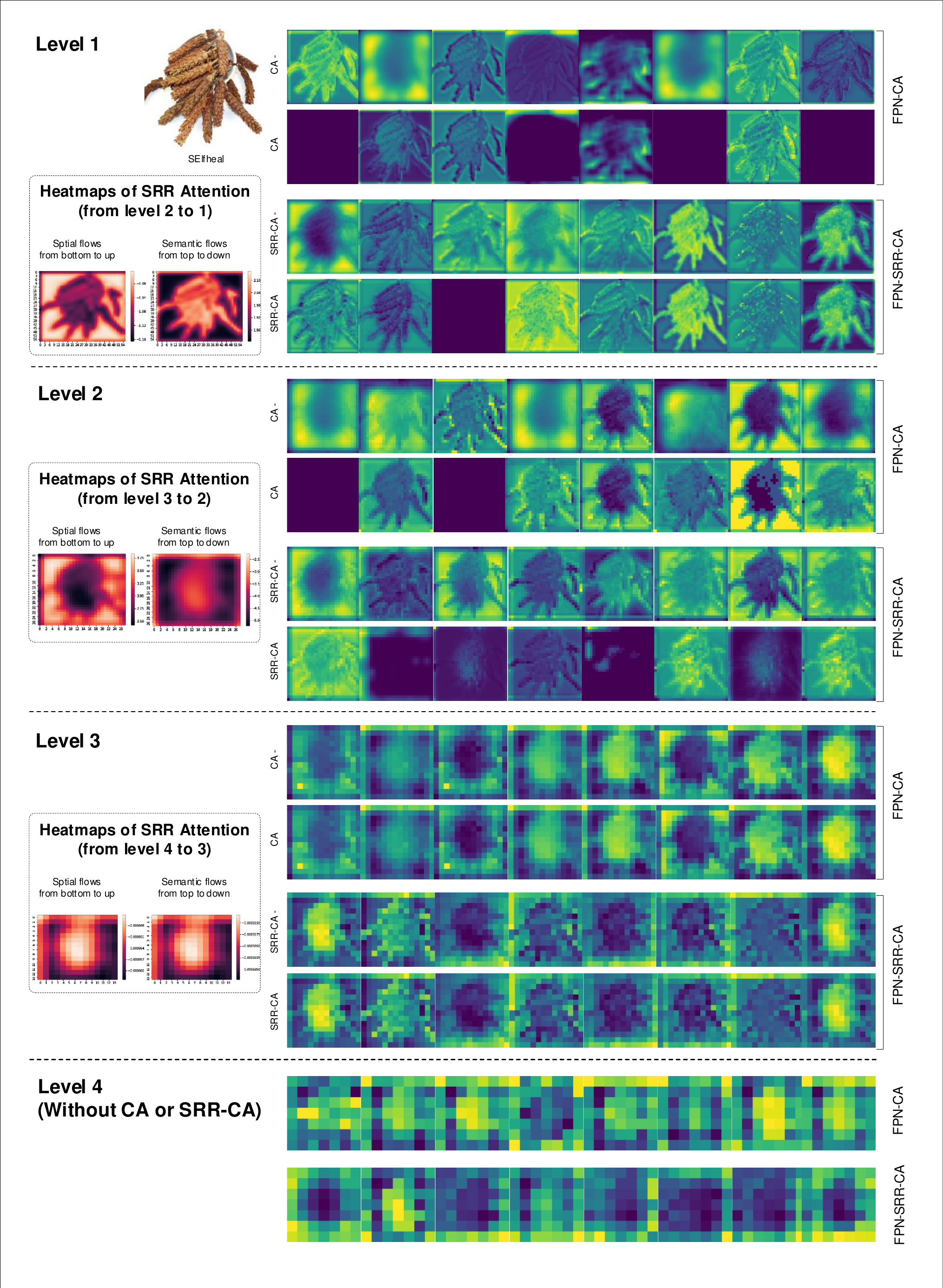}
\caption{\textbf{SElfheal: }intermediate features on various channels from 4 levels and their heatmaps of SRR Attention. For each block (excepted for level 4), the \textbf{top (-)} are not re-scaled by attention and the \textbf{bottom} refers features reweighted on the corresponding channel.}
\label{fig_feature52}
\end{figure*}

\begin{figure*}[t]
\centering
\includegraphics[scale=0.6]{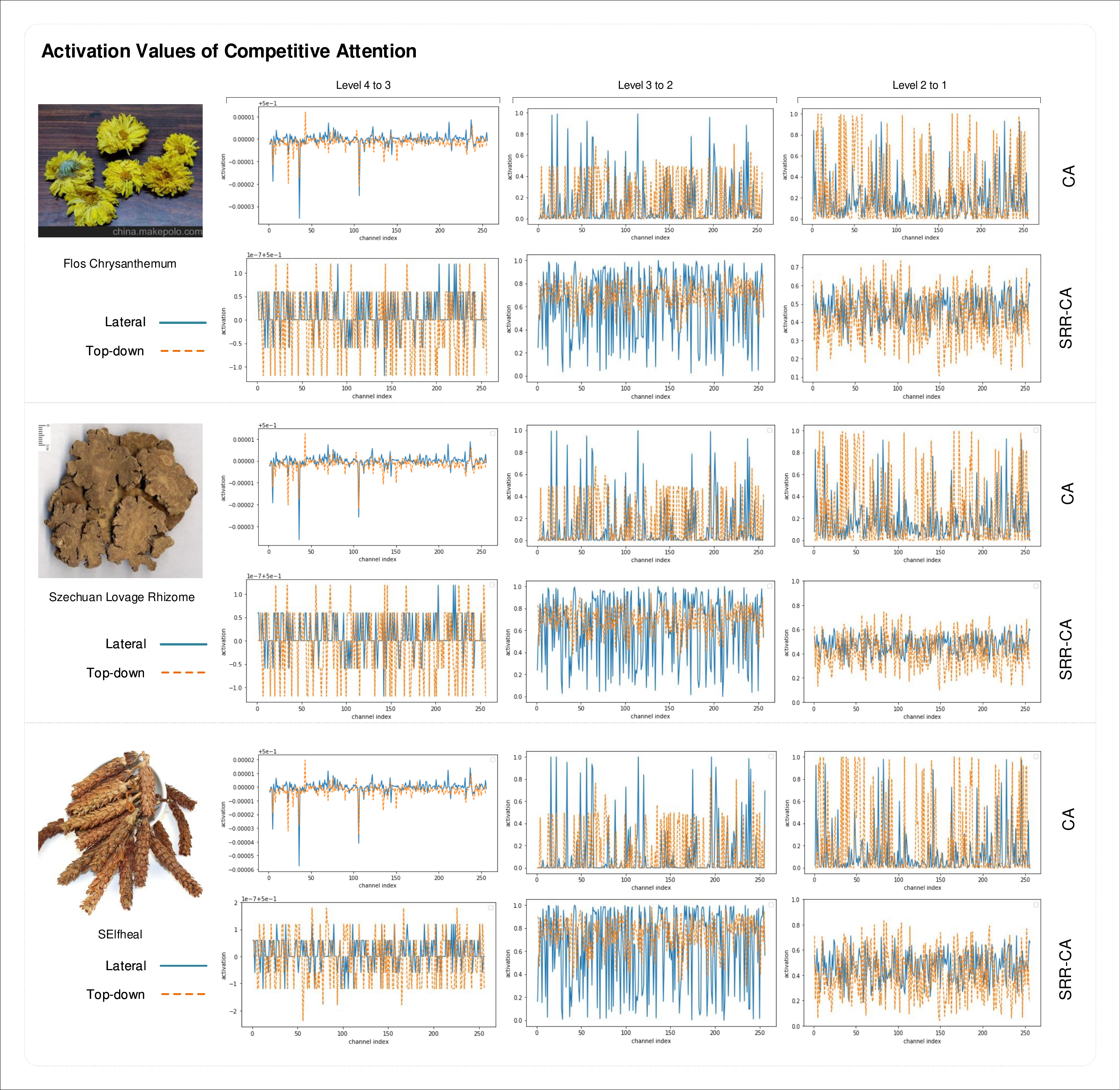}
\caption{Outputs of Competitive Attention on models FPN-CA/SRR-CA. (solid lines for bottom-up pathway via lateral connections and dotted lines for top-down pathway)}
\label{fig_se_statics}
\end{figure*}
In this section, we extract the intermediate features from our models FPN-CA/SRR-CA-18 with ResNet-18 as backbone networks. We define layers producing output maps of same size as one pyramid level and the features extracted from the last layer of various levels are shown in Fig. \ref{fig_feature6}-\ref{fig_feature52} for three examples. Additionally, we statistics activation values of competitve attention from two pathway in the process of merging features and their spatial attention heatmaps.\\
\indent
By observing intuitively, we can obviously see that the informathons of some features on many channels are suppressed, either re-scaling with a small weight or retaining more local features, and with this adjustment models can get better performance, which does confirm our inference in the section 4.2 that the fusion method of original FPN will lead to redundancies in feature maps. That is also one of the motivations for us to propose attention mechanism. Moreover,
the attentional regions can be apparently seen such as serrated petals shape of Flos Chrysanthemum in Fig. \ref{fig_feature6} and we can also see some fuzzy features are recalibrated spatially and presented more clearly. \\
\indent
The aforementioned changes always occur in the low-level of networks for both FPN-CA and FPN-SRR-CA and features from high-level have not been adjusted too much, shown in level 3 of Fig. \ref{fig_feature6} - \ref{fig_feature52}, which can be verified by the activation values statistics in Fig. \ref{fig_se_statics}. The activation values of level 4 to 3 are always kept at about 0.5 and
fluctuate sightly, for the reason that the features of spatial and semantics flows before merging are extracted from the deep layers, which are adjusted enough. However, on the low-level, features from sematics flows represent more vigorously and the others from spatial flow represent more sparsely, which reflects that there is high information density on the semantic flows, which is more benifical to classifying. Furthermore, a majority of features from spatial flows with weak ability of classification are redundant and suppressed, and only a small part of features are selected to make the supplement for semantic flows.\\
\indent
Compared with activation values of competitive attention of FPN-CA, the features on various channels of FPN-SRR-CA are less suppressed. We infer that SRR modules contribute to restoring the spatial informations for misaligned features, which results in higher information density of semantic flow, hence its representation are more vigorous (activation values of CA are almost non-zero), and this situation reflects the SRR-CA module will be more cautious when reducing redundancies of feature maps.\\
\indent
As shown in heatmaps of SRR attention modules, we can see that the attention outputs of different regions are obviously distinguishing and the absolute values of target activation are usually bigger. Howerver, for the examples of Flos Chrysanthemum in Fig. \ref{fig_feature6} of appendix and Unibract Fritillary Bulb in Fig. 5 of main text, we can see the SRR attention focus more on the background on level 1 and we infer that the activation values of SRR attention are closely related to the original images, especially for low-level of networks. The low-level features can highly restore the original images and are more sensitive to colors. Therefore, due to the dark colors of background, the absolute values of backgound activation are bigger than target. Despite all this, SRR attention has played a role in distinguishing from different regions and recalibrated the misaligned features.

\end{document}